\pdfoutput=1
\documentclass[11pt]{article}

\usepackage{ACL2023}
\usepackage{times}
\usepackage{latexsym}

\usepackage[T1]{fontenc}
\usepackage[utf8]{inputenc}

\usepackage{microtype}

\usepackage{inconsolata}
\usepackage{graphicx} 
\usepackage{booktabs}
\usepackage{multicol} % Supports multiple columns
\usepackage{tabularx} % Supports adjustable-width tables
\usepackage{caption}  % For table captions
\usepackage{amsmath} 
\usepackage{multirow}

\title{SSMLoRA: Enhancing Low-Rank Adaptation with State Space Model}
\author{Jiayang Yu$^{1}$, Yihang Zhang$^{1}$, Bin Wang$^1$, Peiqin Lin$^{2,3}$, \\
  \textbf{Yongkang Liu$^{1}$, Shi Feng$^{1}$\footnotemark[1]} \\
        $^1$Northeastern University, China;
        $^2$CIS, LMU Munich, Germany \\
        $^3$Munich Center for Machine Learning (MCML), Germany \\
        \texttt{20221251@stu.neu.edu.cn,misonsky@163.com} \\
        \texttt{fengshi@cse.neu.edu.cn }
}

\begin{document}
\maketitle
\begin{abstract}
\renewcommand{\thefootnote}{\fnsymbol{footnote}} %将脚注符号设置为fnsymbol类型，即特殊符号表示
\footnotetext[1]{Corresponding Author} %对应脚注[2]

Fine-tuning is a key approach for adapting language models to specific downstream tasks, but updating all model parameters becomes impractical as model sizes increase.
Parameter-Efficient Fine-Tuning (PEFT) methods, such as Low-Rank Adaptation (LoRA), address this challenge by introducing additional adaptation parameters into pre-trained weight matrices.
However, LoRA's performance varies across different insertion points within the model, highlighting potential parameter inefficiency due to unnecessary insertions. To this end, we propose SSMLoRA (\textbf{S}tate \textbf{S}pace \textbf{M}odel \textbf{L}ow-\textbf{R}ank \textbf{A}daptation), an extension of LoRA that incorporates a State Space Model (SSM) to interconnect low-rank matrices. SSMLoRA ensures that performance is maintained even with sparser insertions. SSMLoRA allows the model to not only map inputs to a low-rank space for better feature extraction but also leverage the computations from the previous low-rank space. Our method achieves comparable performance to LoRA on the General Language Understanding Evaluation (GLUE) benchmark while using only half the parameters. Additionally, due to its structure, SSMLoRA shows promise in handling tasks with longer input sequences.You can find our code here:\url{https://github.com/yuhkalhic/SSMLoRA}

\end{abstract}

\section{Introduction}

Fine-tuning \cite{park-lee-2021-finetuning} is a technique aimed at enhancing model performance by adjusting the parameters of a pretrained model on specific task data. This approach effectively leverages the knowledge accumulated through large-scale training, accelerating the model's adaptation to new tasks and improving results. However, fine-tuning requires substantial computational resources and is prone to overfitting when applied to small datasets. These limitations have prompted researchers to explore more efficient parameter adjustment strategies.

To address resource consumption and overfitting challenges in fine-tuning, researchers have proposed diverse strategies. \citet{liu2024hifthierarchicalparameterfinetuning} introduced an end-to-end hierarchical fine-tuning approach to mitigate memory constraints associated with full-parameter fine-tuning. Concurrently, a significant research trend has emerged focusing on Parameter-Efficient Fine-Tuning (PEFT) \cite{houlsby2019parameterefficienttransferlearningnlp} techniques, which are a set of methods designed to optimize parameter updates. PEFT updates only a small subset of the model's parameters while keeping the majority of pretrained parameters unchanged. LoRA \cite{hu2022lora}, one of the most commonly used PEFT techniques, introduces low-rank matrices to adjust specific weights in the pretrained model instead of updating all parameters comprehensively. This allows the model to retain most of its structure while adapting to downstream tasks, particularly for transformer-based deep language models. Following the successful application of LoRA, various extensions have emerged to further enhance its adaptability and efficiency. These methods aim to improve the model's capability to perform well in resource-constrained environments.

Despite its success, LoRA and its variants still face certain challenges. SoRA \cite{ding2023sparse} has indicated that LoRA may lead to unnecessary parameter overhead. More efficient methods, such as pruning certain low-rank matrices and using smaller scaling factors to reduce parameter usage, were employed. Moreover, compared to full fine-tuning, parameter-efficient methods tend to have limitations in capturing information, exacerbating the inherent difficulty that transformer-based models face when processing long texts. This leads to the observation that, for many tasks, the number of parameters used by LoRA is excessive, while for tasks involving long-text processing in large language models, it is insufficient.

To address this issue, we propose State Space Model Low-Rank Adaptation (SSMLoRA), a novel technique that builds on the state-space equation framework to insert sparse low-rank matrices. We introduce interval layers and apply our technique selectively to the query and value matrices within the attention mechanism, as this design significantly reduces parameter overhead. Additionally, we incorporate a State Space Model to connect low-rank matrices inserted into the same type of neural network layers. State Space Model, which excels at handling long-text tasks, is based on state transitions but differs from traditional RNNs by enabling parallel computation through FFT-based transformations of the state transitions. This enhancement facilitates parallel training in our approach. Experimental results demonstrate that our technique performs well when applied to both linear and convolutional layers. Compared to LoRA, SSMLoRA achieves superior performance across a wide range of tasks while requiring only half of the parameters. Furthermore, it excels in long-text and certain other specific tasks. We believe that this represents a new direction in the exploration of LoRA-derived techniques, adopting a more holistic and macro-level perspective.

% Fine-tuning是一种通过在特定任务数据上调整预训练模型参数以提升模型性能的技术。这一方法有效地利用了大规模训练所积累的知识，加速了模型在新任务上的适应性，并改善了结果。然而，fine-tuning需要大量计算资源，并且在小数据集上容易导致过拟合，这些限制促使研究者探索更为高效的参数调整策略。

%为应对fine-tuning中资源消耗和过拟合问题，具有许多思路。\citet{liu2024hifthierarchicalparameterfinetuning}提出了一种end-to-end hierarchical fine-tuning strategy降低全参微调所需的显存；而更多研究者提出了参数高效微调（PEFT）技术，这是一系列高效的参数调整方法。PEFT仅更新模型中的一小部分参数，而大多数预训练参数保持不变。LoRA是最为常用的一种参数高效微调技术，引入低秩矩阵来调整预训练模型中的特定权重，而非全面更新所有参数，能够保持模型大部分结构不变，适应下游任务，适用于当前基于transformer结构的深层的大语言模型。在 LoRA 技术成功应用后，涌现了多种延伸版本，旨在增强其适应性和效率。这些方法旨在增强与模型的适应性和效率，并且使得模型在环境配置较差时也可以顺利训练。

%然而LoRA以及各种衍生方法仍然存在一些问题。一些研究例如SoRA微调技术表明了LoRA微调技术存在着一定的参数浪费，可以采用更加节省的方式，它提出了可以裁剪一些低秩矩阵设置更小的缩放因子。除此之外，参数高效的微调方法相比全参数微调会在信息捕获能力方面有所不足，这个会进一步加大基于transformer架构的模型本身具有的处理长文本能力不足的问题。因此我们可以得出这样的一个观察，针对于大部分任务，LoRA技术使用的参数量是过剩的，但在微调大语言模型解决长文本任务时的又是不足的。因此我们希望提出了基于状态空间方程的稀疏化插入低质矩阵的技术，称之为SSMLoRA。

%我们首先设置了一些间隔层，并且对于注意力机制模块，我们的技术只在效果最好的query和value矩阵使用，这样的设计显著降低了参数量。我们还引入了状态空间模型来连接插入同一个类型的神经网络层的低秩矩阵，这是一种善于处理长文本的模型结构，但区别于传统RNN，它可以将前后状态通过FFT技术变换而解决了无法并行计算的问题。我们的技术在作用于线性层和卷积层时均有良好的表现，并且能够在仅添加LoRA技术一半参数量的情况下，具有和LoRA技术相似的表现，并且在长文本的任务和一些其它的任务中有更优异的性能。

\section{Related Works}

\subsection{PEFT}
Parameter-Efficient Fine-Tuning (PEFT) \cite{houlsby2019parameterefficienttransferlearningnlp} aims to reduce the computational and storage costs associated with fine-tuning large language models (LLMs). By tuning only a small subset of additional parameters while freezing the majority of pretrained model parameters, PEFT effectively mitigates the catastrophic forgetting issue often encountered in full fine-tuning. Moreover, PEFT outperforms traditional fine-tuning methods in low-data environments, demonstrating superior generalization capability. This technique is not limited to natural language processing tasks but extends to computer vision and audio domains, significantly enhancing the model's flexibility and adaptability. LoRA, as one of the most widely adopted PEFT techniques, adjusts specific pretrained model weights by introducing low-rank matrices, rather than updating all parameters, making it well-suited to downstream tasks. LoRA has shown exceptional adaptability and efficiency in large language models, allowing them to swiftly adapt to new tasks while preserving most of their original structure. The success of LoRA has sparked numerous further studies in the field.

AdaLoRA \cite{zhang2023adaloraadaptivebudgetallocation}, for instance, introduces orthogonal regularization to ensure that the low-rank projection matrices comply with Singular Value Decomposition (SVD), thus avoiding the reliance on incremental updates, albeit at the cost of increased computational complexity. QLoRA \cite{dettmers2023qloraefficientfinetuningquantized} improves computational efficiency and reduces resource consumption through dynamic quantization and advanced strategies, though it may potentially impact model accuracy. DoRA \cite{liu2024dora} decomposes pretrained weights into magnitude and direction components, utilizing LoRA for directional updates, reducing trainable parameters and enhancing fine-tuning performance, though its complexity and dependence on data quality may limit its effectiveness. Vera \cite{kopiczko2024veravectorbasedrandommatrix}, on the other hand, reduces parameter overhead by sharing low-rank matrices and learning small-scale vectors, while maintaining performance, though the model's complexity and hyperparameter tuning requirements might hinder overall effectiveness.

\subsection{State Space Model}
State Space Model (SSM) are mathematical models used to describe dynamic systems, comprising an input sequence \( x(t) \), a latent state representation, and an output sequence \( y(t) \). The primary objective of an SSM is to predict future states based on the current input and previous states. In recent years, researchers have made significant progress in applying SSMs to neural network architectures. Compared to traditional Recurrent Neural Networks (RNNs), SSM-based models are capable of resolving the issue of sequential processing bottlenecks, thus significantly improving training efficiency. Models based on SSMs, such as HIPPO (Hierarchy of Integrators with Projection and Partial Orthogonalization \cite{hippo}, have been shown to capture complex temporal dependencies more effectively. S4 (Structured State Space Sequence Model) \cite{gu2022efficiently} leveraged SSM architecture to outperform transformers and other architectures in handling ultra-long sequences. Mamba \cite{mamba}, an extension of SSM, introduced a selection mechanism that dynamically adjusts the model based on input, thereby improving adaptability. However, despite these advantages, the generalization capabilities of SSMs across various domains remain to be fully validated. To address this, we propose incorporating SSM into fine-tuning methodologies. By improving the existing LoRA (Low-Rank Adaptation) technique, we aim to enhance fine-tuning performance on specific tasks while preserving the Transformer architecture. This leads to our proposed method, SSMLoRA, which combines modified SSM equations with LoRA. Compared to traditional LoRA and its variants, SSMLoRA strengthens the connections between inserted low-rank matrices, improving data matching capabilities and achieving comparable or superior performance across most datasets.

\subsection{Sparsification Methods}
Sparsification methods aim to reduce computational complexity and memory requirements by decreasing the number of model parameters while maintaining performance as much as possible. These techniques include weight pruning, structured sparsity, dynamic sparse training, and sparse regularization. \citet{Wen_NIPS2016} proposed Structured Sparsity Learning (SSL), a technique capable of learning efficient and compact structures from large deep neural networks, significantly enhancing model speed and performance. \citet{anwar2015structuredpruningdeepconvolutional} optimized convolutional neural networks through structured pruning, effectively reducing overall memory demand. \citet{xie2018igcv2interleavedstructuredsparse} proposed IGCV2, which applied interleaved structured sparsity to convolutional layers. These studies demonstrate the effectiveness and generalizability of employing sparser structures. In our work, we aim to improve the parameter utilization of low-rank matrices while significantly reducing memory requirements and improving computational efficiency without sacrificing performance. This provides vital support for subsequent model optimization and deployment.

\section{SSMLoRA}
\begin{figure*}[h] % 开始插入图片
    \centering % 图片居中
    \includegraphics[width=\linewidth]{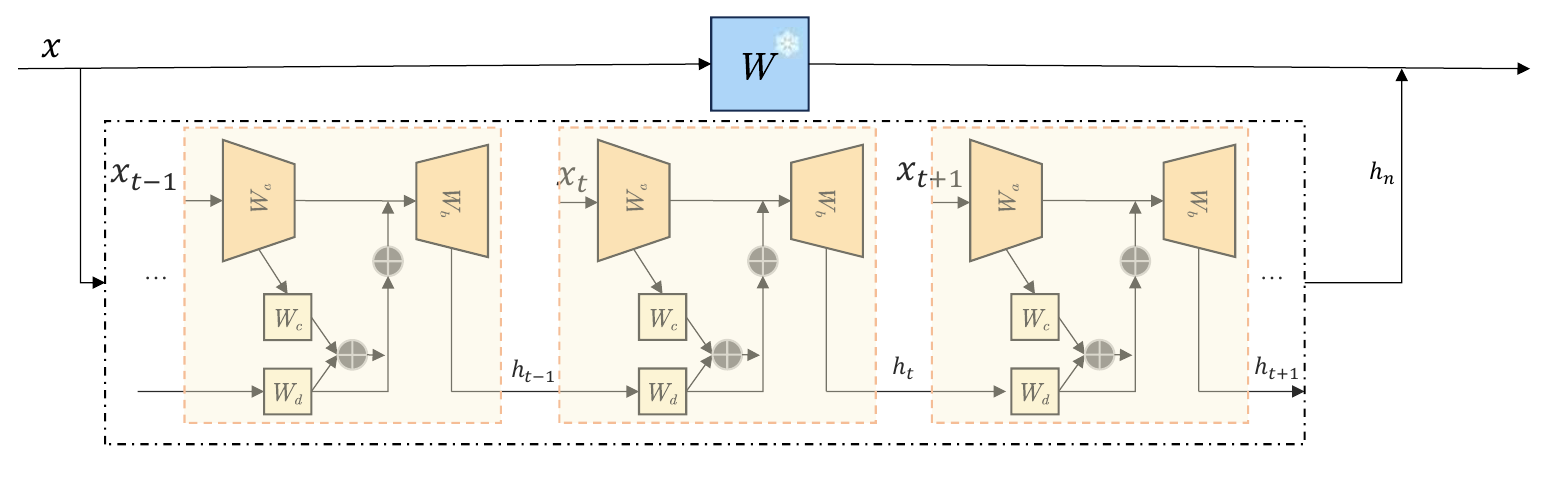} % 插入图片并设置宽度
    \caption{This image illustrates how our technique is applied to the original pre-trained model. In the lower part of the image, the dashed box contains three orange blocks, each representing a Time module. \( W \) refers to a layer from the original pre-trained matrix, which could be a linear layer, convolutional layer, or query, key, and value matrices from the self-attention mechanism. When SSMLoRA fine-tuning is applied to a specific type of neural network layer, these layers are connected along a time axis. Each layer is assigned a position on this axis and linked to a corresponding Time module that adjusts its output. Here, \( t \) denotes the position in the time axis, and \( n \) represents the maximum sequence length.}
\end{figure*}

%这张图片主要展示了我们的技术作用于注意力机制时的插入方式。主要分为几个核心部分。其中图片最右侧白底的矩形表示预训练模型，我们只显示出了多头注意力机制并用橙色方块标注了当前是哪一个层，其余的神经网络层被我们忽略。而完整的矩形表示采用SSMLoRA技术微调后的模型。指向黄色方块的黑色箭头表明了在这个层我们的技术作用于注意力机制的哪个部分。右上方则表示了具体的连接方式，为方便表示只展示出一个head。右下方则是代表了SSMLoRA技术的核心组件Time module的组织形式。

%这张图片主要展示了我们的技术作用于原始预训练模型的方式。图片靠下的虚线方块中有三个橙色方块，我们称每个方块为一个Time module。\( W_l \)为原始预训练矩阵的一个层，可以是线性层，卷积层，或者是自注意力机制中的query、key、value矩阵。当对某种类型的神经网络层应用SSMLoRA微调技术后，它们会被一条时间轴连接，被作用的层都会有其在对应时间轴的位置编码，都会被分配到一个Time module调整其输出的行为。其中t表示为状态的位置，n表示为最大的句子长度

\subsection{Time Module}
In the original LoRA technique, the core concept involves introducing two new projection matrices, \( W_a \) and \( W_b \). Let \( W_0 \) represent the weight matrix of the pre-trained model at a given layer, and let the input vector \( x \) have a dimensionality of \( d \). After scaling, the input vector, denoted as \( x_{\text{new}} \), is mapped to a lower-dimensional space with a rank \( r \). The projection matrices \( W_a \) and \( W_b \), with dimensions \( d \times r \) and \( r \times d \) respectively, project \( x \) to a lower dimension and then back to its original dimension, ensuring that the output dimensionality matches the input. Throughout this process, the parameters of the pre-trained model remain frozen. As a result, compared to training the original \( W_0 \), the adaptation to downstream tasks is achieved by training only the small matrices \( W_a \) and \( W_b \), significantly reducing computational costs. The model is able to fit downstream tasks within the low-rank space. Therefore, after introducing these two matrices, the model's output can be expressed as:
\begin{equation}
y = x \times W_0 + x \times W_a \times W_b
\end{equation}

In our method, we introduce a new module called the Time Module, represented by each orange square in the dashed box in Figure 1. Each Time Module contains the two projection matrices \( W_a \) and \( W_b \) from the original LoRA technique, as well as two additional \( r \times r \) matrices, \( W_c \) and \( W_d \), whose functions will be described later. In each Time Module, we receive the input \( x \) along with the time state \( h_{t-1} \) calculated by the previous module. The Time Module computes the output to adjust the behavior of the pre-trained model and updates the state to \( h_t \), which is then passed to the next Time Module. Our modules can be inserted into various neural network layers, such as linear layers, convolutional layers, and multi-head attention mechanisms, and operate independently of the pre-trained model.

%在原始的LoRA技术中，最核心的逻辑在于引入两个新的缩放矩阵\( W_a \)和\( W_b \)。在这个部分我们设该层的预训练模型该层的权重为\( W_0 \)，输入向量为x，它的维度为d。经过缩放的输入向量称之为x_{\text{new}}，它的维度为缩放的因子r。那么可以得出，缩放矩阵\( W_a \)和\( W_b \)的维度依次为d\times r和r\times d，它们将输入x放缩到一个较低的维度，变换为x_{\text{new}}，然后再放缩回原始维度，以确保输出与输入的维度一致。在这一过程中，原始的预训练模型的所有参数始终保持冻结状态，这样相比于训练原始的\( W_0 \)，仅通过训练少量的 \( W_a \)和 \( W_b \)对实现了对下游任务的适配能够减少许多的计算量，而且在低秩空间中就足以让模型拟合下游的任务。因此在这个位置，引入了两个新的矩阵后，模型的输出可以表示如下：
% \begin{equation}
% y = x \times W_0 + x \times W_a \times W_b
% \end{equation}
% 在我们的方法中，我们引入了一个新的模块称之为Time module，对应于Figure 1下方虚线框内的每一个橙色正方形。对于每个这样的模块，包含了原始LoRA技术的两个缩放矩阵\( W_a \)和\( W_b \)，以及两个维度为r \times r的矩阵\( W_c \)和\( W_d \)，它们的作用将会在后文详细介绍。在每个Time module中，我们会接收输入x和由前一个模块计算出的时间状态\( h_t-1 \)。我们会计算出输出调整预训练模型的行为并且更新状态为\( h_t \)并将其传递给下一个Time module。可以看到，我们的模块支持插入到线性层、卷积层、多头注意力等多种神经网络层，与预训练模型相互独立。

\subsection{TimeAxis Update}
In SSMLoRA, the core innovation, similar to LoRA, lies in the integration of the Time Module with specific pre-trained layers. Before discussing how the Time Module propagates forward, it is essential to understand how it retrieves the state computed by the previous Time Module and updates it. In the context of State Space Model, the matrix \( W_c \) within the Time Module is commonly referred to as the state matrix, while \( W_d \) is known as the control matrix. These matrices are used to combine the previous state and update the current state in the time cycle, which is particularly effective for extracting information from long sequences. Building on this concept, we follow the principles of the S4 model, utilizing these two matrices to update the state from the previous time step \( h_{t-1} \) to the current state \( h_t \), and propagate it forward. Specifically, the process is as follows:

First, we compute the derivative of \( h_{t-1} \) at the current position using the following formula, which is based on the classical State Space Model (SSM) equations, producing the derivative of the current state:
\begin{equation}
h_t' = h_t \times W_c + x_{\text{new}} \times W_d
\end{equation}

This equation corresponds to the first addition operation in the Time Module structure shown in Figure 1. Next, unlike the S4 or SSM methods, we use a Taylor expansion to calculate the increment for the next time step, which corresponds to the second addition operation in Figure 1:
\begin{equation}
h_{t+1} = h_t' + h_t
\end{equation}

Thus, the complete computation process can be expressed as follows:
\begin{equation}
h_{t+1} = h_t \times W_c + x_{\text{new}} \times W_d + h_t
\end{equation}

It is important to note that for the Time Module at the beginning of the time axis, \( h_{t-1} \) is initialized to a zero vector of the same dimensionality as \( x_{\text{new}} \). This initialization helps maintain numerical stability and ensures that the model is not adversely affected during the early stages of training.

Using this method, we compute the current time state \( h_t \) based on the previous state \( h_{t-1} \). Our approach, which approximates the current time value using a Taylor expansion at time step \( t \), is motivated by several considerations. First, we can discretize the state directly, transforming the continuous function into a discrete representation corresponding to the spatial position of the Time Module. This avoids the need for discretizing \( W_c \) and \( W_d \), as done in S4, thus reducing computational overhead. Discretization ensures that the model retains both precision and stability when processing sequential data, and by appropriately controlling the time steps, the model's adaptability to long sequences is enhanced. Moreover, compared to the original SSM or S4 models, we reduce parameter usage by directly utilizing the current time value \( h_{t+1} \) to adjust the output of the pre-trained model. Furthermore, leveraging the S4 model, we can apply FFT-based matrix transformations, enabling our method to overcome the parallelization limitations typically encountered in RNNs.

In our method, \( W_c \) and \( W_d \) are new parameters that participate in training, while \( h_t \) is detached from the computational graph and does not participate in training. Therefore, the state update is controlled solely by \( W_c \), \( W_d \), and the input \( x \).

% 在 SSMLoRA 技术中，其核心创新与lora类似，是将Time module与特定预训练层结合，在讨论Time module如何前向传播之前，探讨一下它是如何获取前一个Time module计算的状态的值并且如何并且更新它是十分重要的。值得说明的是，在状态空间模型的定义中，Time module模块内的\(W_c\)矩阵普遍被称为状态矩阵，而\(W_d\)矩阵一般被称为控制矩阵，它们用于结合前一个时刻的状态来更新当前时间周期的状态，这样做对于长序列的数据具有很好的提取能力。而我们在此基础上参照了S4模型的原理，利用这两个矩阵来更新时间轴的前一个位置的状态\( h_t-1 \)获取当前的状态\( h_t \)并向前传递，具体而言是这样：
%先通过以下的公式来求出\( h_t-1 \)在当前位置的导数，这个公式是基于SSM模型中的经典公式，计算出了当前状态求导之后的结果：
% \begin{equation}
% h_t' = h_t \times W_c + x_{\text{new}} \times W_d
% \end{equation}

% 这个公式对应于Figure 1的Time module结构中第一个加法操作示例以前的位置。接着，区别于S4或者SSM的方法，我们采用泰勒展开计算出下一个时间步的增量，该过程同样可以在Figure1中看到，对应于第二个加法步骤，即：

% \begin{equation}
% h_{t+1} = h_t' + h_t
% \end{equation}

% 因此该过程的完整计算步骤可以被表示成以下的公式：

% \begin{equation}
% h_{t+1} = h_t \times W_c + x_{\text{new}} \times W_d + h_t
% \end{equation}

%需要注意的是，对于位于时间轴最前端的Time module组件，我们会将\( h_t-1 \)采用零初始化为和x_{\text{new}}同维度向量的方式，这样的操作有助于数值稳定性，在训练初期不会对模型进行影响。
% 通过这种方法，我们可以从获取的时间状态\( h_t-1 \)计算出当前的时间状态\( h_t \)。我们的通过\( h_t-1 \)在t处的近似泰勒展开计算出当前的时间值的方法源于几点考虑。首先我们可以直接将状态进行离散化，直接将连续函数转化为可以代表Time module空间位置的位置，无需效仿S4对\(W_c\)和\(W_d\)进行离散化操作，降低了计算量。离散化的实施确保了模型在处理时间序列数据时，能够保持计算的精确性和稳定性，并通过合理控制时间步，增强了模型在长时间序列处理中的适应能力。除此之外，相比于原始的SSM或者S4，我们节约了参数的使用，直接采用当前的时间值\( h_t+1 \)来调整原始模型的输出。并且基于S4模型，我们可以进一步利用FFT进行矩阵变换，使我们的方法突破类似RNN无法进行并行训练的限制。

%在我们的方法中，\(W_c\)和\(W_d\)作为新的参数参与训练，而\( h_t \)则被从计算图中切断不参与训练。因此它的更新仅由\(W_c\)和\(W_d\)和输入x进行控制。

\subsection{Tuned Model Forward}

As discussed earlier, our proposed SSMLoRA technique introduces a state transition mechanism based on the State Space Model (SSM) architecture, specifically by adding two matrices. The detailed architecture is shown in Figure 1, where we illustrate the structure of the Time Module, which forms a timeline based on the SSM model, represented by the dashed section on the right side of the figure. At each Time Module, the current state \( h_t \) is retrieved, followed by the computation of the output \( y \) and the next state \( h_{t+1} \). This mechanism is introduced into the pre-trained model to dynamically adjust the feature representation of the input \( x \), enhancing the model’s performance on tasks that are highly dependent on temporal information.

First, we clarify the origin of the variable \( x_{\text{new}} \), used throughout the previous sections, which is obtained by projecting the input \( x \) into a low-rank matrix:
\begin{equation}
x_{\text{new}} = x \times W_a
\end{equation}
The dimensionality of the projected low-rank matrix \( x_{\text{new}} \) becomes \( r \), which matches the dimensionality of the current state value \( h_{t+1} \) calculated through the process described in Section 3.2, allowing direct computation. However, for numerical stability, we normalize \( h_{t+1} \) before performing further calculations, as outlined in the following equations:

\begin{equation}
\text{min\_val} = \min(h_{t+1})
\end{equation}
\begin{equation}
\text{max\_val} = \max(h_{t+1})
\end{equation}
\begin{equation}
h_{t+1}^{\text{norm}} = \frac{h_{t+1} - \text{min\_val}}{\text{max\_val} - \text{min\_val} + \epsilon}
\end{equation}

Next, before restoring \( x \) to its original dimensionality, we adjust \( x \) using the normalized \( h_{t+1} \). Therefore, the output of the Time Module is given by:
\begin{equation}
y = x_{\text{new}} + h_{t+1}^{\text{norm}} \times W_b
\end{equation}

In summary, the output of the original pre-trained weight matrix can be expanded as follows, with our technique introducing additional adjustments using fewer parameters and operations than LoRA:
\begin{equation}
y = x \times W_0 + (x \times W_a + h_{t+1}^{\text{norm}}) \times W_b
\end{equation}

Since each Time Module is influenced by the preceding Time Module along the timeline, we design the Time Modules to be placed on different timelines depending on the type of neural network layer they are inserted into, preventing mutual interference. For instance, in the self-attention module of the model, if a timeline is set, separate timelines will be established for the query, key, and value matrices to prevent temporal misalignment. Therefore, the example shown in Figure 1 demonstrates the insertion into a single neural network layer, rather than a complete example across the entire model.

% 在这个部分，将会详细介绍每个Time module插入预训练模型后是如何进行前向传播的。
% 在我们提出的 SSMLoRA 技术中，正如前文探讨的，我们引入了基于状态空间模型（SSM）架构的状态转换机制，具体来说是增加了两个矩阵。具体的架构图由Figure 1所示，我们引入了Time module的结构，组合成了一个基于SSM模型的时间轴，也就是图右侧的虚线部分。在每个Time module处，获取当前的状态\( h_t \)，接着计算输出y的和下一个状态\( h_t+1 \) 。在预训练模型中引入由该机制旨在动态调整输入 \( x \) 的特征表示，增强模型在处理时间依赖性强的任务中的表现。首先我们先补充前文一直使用的变量\( x_new \)的来源，即我们先将x投影为低秩矩阵：
% \[
% \( x_new \) = x \times W_a
% \]
%投影后的低秩矩阵\( x_new \)的维度变为了r，具有与3.2描述的过程计算出的当前状态值\( h_t+1 \)相同的维度，可以直接的进行计算。但在计算前为了数值的稳定性，对\( h_t+1 \)进行归一化操作，过程如下三个公式所示：

% \begin{equation}
% \text{min\_val} = \min(h_{t+1})
% \end{equation}
% \begin{equation}
% \text{max\_val} = \max(h_{t+1})
% \end{equation}
% \begin{equation}
% h_{t+1}^{\text{norm}} = \frac{h_{t+1} - \text{min\_val}}{\text{max\_val} - \text{min\_val} + \epsilon}
% \end{equation}

%接着在将x还原回原始维度前，使用归一化后的\( h_t+1 \)调整x，因此Time module的输出应该为
% \[
% y = (\( x_new \) + h_{t+1}^{\text{norm}} \times \( W_b \)
% \]

%综合以上的讨论内容，我们的将原始的预训练权重矩阵的输出扩展为如下表达式，我们的技术可以被认为在LoRA技术的基础上采用少量参数和少量运算进一步的进行调整：

% \[
% y = x \times W_0 + (x \times W_a + h_t+1) \times W_b
% \]

% 由于每个 Time module 需要受到位于时间轴前端的 Time module 的影响，因此在我们的设计中，我们针对插入到不同类型神经网络层的 Time module 设置于不同的时间轴，防止相互影响。例如，对于整个模型的自注意力模块，如果设置了相应的时间轴，则会在 query、key、value 三个矩阵上分别设置三个不同的时间轴，以防止出现时序错乱。也就是说，Figure 1所展示的示例图展示的是只针对于一种神经网络层，而非完整的示例。

\subsection{Sparse Insertion}

As emphasized earlier, our method applies a sparsification approach to the attention mechanism. Research on LoRA has shown that, for large language models based on the transformer architecture, the best results are achieved when both the query and value matrices are adjusted simultaneously. Therefore, in this section, we adopt a more efficient strategy by applying the Time module only to the query and value matrices in the attention mechanism.

We further found that sparsifying the full structure does not lead to any loss of accuracy. Specifically, in each encoder's attention module, SSMLoRA operates on only one of the query or value matrices at a time, meaning one matrix is "activated" for adjustment. We recommend alternating sparse insertions in a regular pattern, which is extremely easy to implement. As illustrated in Figure 1, if $W$ represents the query matrix of the $l$-th layer's attention mechanism, the query matrices in layers $l+1$ and $l-1$ are excluded from the time axis, while the value matrix follows a separate time axis.

For other non-multihead attention components, our method essentially replaces LoRA with SSMLoRA, maintaining a compact parameter configuration. This approach, as demonstrated in Section 4.1, outperforms other fine-tuning methods, achieving better results with less than 80\% of the parameters used by LoRA.

%正如在前文强调的，我们的方法采用稀疏化的方法作用在注意力机制上。对于普遍的基于transformer架构的大语言模型，由LoRA技术的研究我们可以知道，同时作用于矩阵query和value，调整它们的输出时，会得到最好的效果。因此我们在这个部分采用更为节省的策略，针对注意力机制，我们的Time module只作用于query和value。我们进一步发现将完整的结构进行进一步稀疏不会损失精度，具体而言的过程是这样的：对于每一个encoder的attention模块，SSMLoRA技术只会作用于query和value矩阵其中的一个，也就是会在这两个矩阵中选择一个进行"激活"，使其的输出受到调整。我们推荐采用规律的交替间隔插入，这在实现上极为方便。如Figure 1所示，如果此时的$W$代表的是第l层的自注意力机制中的query矩阵，那么第l+1和第l-1层的自注意力机制中的query矩阵将不存在于时间轴中，而value矩阵将会处于一条与图中不同的时间轴。。而对于其它非多头注意力的位置，我们的技术仅相当于将LoRA微调技术替换为SSMLoRA微调技术，仍然采用较为紧凑的方式。这样的使用方式在4.1部分被证实了效果，与其它的一些微调技术进行对比，它能够以不到LoRA技术80%的参数量而实现更好的效果。

% \subsection{初始化策略}
% 在SSMLoRA模型中，我们引入了四个关键矩阵 $W_a$、$W_b$、$W_c$ 和 $W_d$ 以实现状态转换和特征映射。为确保初始化阶段的稳定性和有效性，我们采用了差异化的初始化策略。此外，尽管时间向量 $h_t$ 主要作为传递量而非独立训练参数，其初始化方法仍值得深入探讨。
% 具体而言，对于矩阵 $W_a$ 和 $W_b$，我们借鉴了LoRA微调技术的初始化方法。$W_a$ 的初始化采用了缩放的高斯分布随机值，通过引入适度的随机性，有效避免了零初始化可能导致的训练收敛缓慢问题。相比之下，$W_b$ 被初始化为零矩阵，以确保初始阶段整体输出不受显著影响，同时允许模型在训练过程中逐步学习有效的映射关系。
% 对于矩阵 $W_c$、$W_d$ 以及时间向量 $h_t$，我们统一采用零初始化策略。根据公式(8)所示，我们主要通过调整偏置项来优化LoRA技术的输出。这种方法旨在训练初期最小化引入额外噪声，随后逐步学习时间状态的权重分布。因此，SSMLoRA在训练初期可视为稀疏化的LoRA变体，这可能导致初始性能略低于标准LoRA。然而，随着训练的深入，模型能够学习不同层低秩矩阵之间的内在联系，从而实现性能的稳步提升，最终超越传统LoRA技术。
% 在第4节的实验中，我们发现SSMLoRA在信息密度较高的短文本匹配、问答任务，以及长文本序列处理任务上均表现出优于LoRA的性能。这一结果充分证明了我们所提出的初始化策略和微调技术的有效性。
\subsection{Initialization Strategy}
In the SSMLoRA model, we introduce four key matrices $W_a$, $W_b$, $W_c$, and $W_d$ to achieve state transitions and feature mapping. To ensure stability and effectiveness during the initialization phase, we adopt a differentiated initialization strategy. Furthermore, although the time vector $h_t$ primarily serves as a transmitted quantity rather than an independently trained parameter, its initialization method still warrants in-depth exploration.

Specifically, for matrices $W_a$ and $W_b$, we draw inspiration from the initialization method of the LoRA fine-tuning technique. The initialization of $W_a$ employs scaled Gaussian-distributed random values, effectively introducing moderate randomness to avoid the potential slow convergence problem associated with zero initialization. In contrast, $W_b$ is initialized as a zero matrix to ensure that the overall output remains unaffected in the initial stage while allowing the model to gradually learn effective mapping relationships during the training process.

For matrices $W_c$, $W_d$, and the time vector $h_t$, we uniformly adopt a zero initialization strategy. As shown in Equation (8), we primarily optimize the output of the LoRA technique by adjusting the bias term. This approach aims to minimize the introduction of additional noise in the early stages of training, subsequently learning the weight distribution of temporal states progressively. Consequently, SSMLoRA can be viewed as a sparsified variant of LoRA in the initial training phase, which may result in slightly lower initial performance compared to standard LoRA. However, as training progresses, the model can learn the intrinsic connections between low-rank matrices across different layers, thereby achieving steady performance improvements and ultimately surpassing traditional LoRA techniques.

In the experiments presented in Section 4, we demonstrate that SSMLoRA outperforms LoRA in tasks involving high-density short text matching, question answering, and long-text sequence processing. These results provide compelling evidence for the effectiveness of our proposed initialization strategy and fine-tuning technique.

\section{Experiments}

% 为验证SSMLoRA的有效性，我们设计并执行了一系列全面的实验。我们的实验内容主要围绕两个核心方向展开：
% (1) 作为一种参数高效方法的综合评估：我们将SSMLoRA与全参数微调以及现有的其他参数高效微调方法进行了多方面的性能对比和分析。这一系列实验旨在探究SSMLoRA在不同任务类型和应用场景下的表现，验证SSMLoRA微调技术作为一种具有潜力能够被广泛运用的技术。我们进一步识别这项新技术擅长的领域，分析并深入理解其性能优势的根源。
% (2) 与LoRA技术进行系统性对比：我们希望强调一下SSMLoRA技术中稀疏化插入这一关键技术要点的效果。鉴于SSMLoRA的可训练参数，尤其是相对于LoRA微调技术的引入的新参数主要受秩(rank)的影响，我们设置了多种r的值进行对比。我们特别关注了SSMLoRA采用最为稀疏的方式插入，在极低参数量条件下与LoRA微调技术的性能对比。这一对比旨在评估SSMLoRA在参数效率方面的优势。
%我们的实验采用了多种基准模型，我们使用了RoBERTa-base，RoBERTa-large，DeBERTaV3-V3-base，GPT-2这四种模型，采用了GLUE和SUPERGLUE两个基准，以及比较了LoRA，fine-tune等多种微调技术，较好的证明了我们的方法的有效性。

\subsection{Datasets} 
We comprehensively evaluate our methodology on two widely-adopted benchmarks: GLUE and SuperGLUE. The GLUE benchmark encompasses diverse tasks, including linguistic acceptability (CoLA; \cite{warstadt2018neural}), sentiment analysis (SST-2; \cite{socher2013recursive}), paraphrase detection (MRPC; \cite{dolan-brockett-2005-automatically}), semantic textual similarity (STS-B; \cite{cer-etal-2017-semeval}), question pair equivalence (QQP), natural language inference (MNLI; \cite{williams-etal-2018-broad}, QNLI), and textual entailment (RTE; \cite{dagan2005pascal}). The SuperGLUE benchmark comprises advanced language understanding tasks: commonsense inference (CB; \cite{de2019commitmentbank}, COPA; \cite{roemmele2011choice}), multi-sentence reading comprehension (MultiRC; \cite{MultiRC2018}), textual entailment (RTE; \cite{dagan2005pascal}), word sense disambiguation (WiC; \cite{pilehvar2019wicwordincontextdatasetevaluating}), coreference resolution (WSC; \cite{}), boolean question answering (BoolQ; \cite{clark2019boolqexploringsurprisingdifficulty}), and reading comprehension (ReCoRD; \cite{zhang2018recordbridginggaphuman}). Additionally, to evaluate our method's efficacy on long-form text processing, generation capabilities, and needle-in-a-haystack scenarios, we conduct experiments on SQuAD \cite{rajpurkar-etal-2016-squad}, NarrativeQA \cite{kočiský2017narrativeqareadingcomprehensionchallenge}, and RACE \cite{lai-etal-2017-race}.

% \textbf{数据集.} 我们主要使用GLUE和SuperGLUE这两个benchmark来全面地评估我们的方法。GLUE包含了CoLA \cite{warstadt2018neural}, SST-2 \cite{socher2013recursive}, MRPC \cite{dolan-brockett-2005-automatically}, STS-B \cite{cer-etal-2017-semeval}, QQP, MNLI \cite{williams-etal-2018-broad}, QNLI, and RTE \cite{dagan2005pascal}等多个数据集；SuperGLUE包含了CB \cite{de2019commitmentbank}, COPA \cite{roemmele2011choice}, MultiRC \cite{MultiRC2018}, RTE \cite{dagan2005pascal}, WiC \cite{pilehvar2019wicwordincontextdatasetevaluating}, WSC \cite{}, BoolQ \cite{clark2019boolqexploringsurprisingdifficulty}, and ReCoRD \cite{zhang2018recordbridginggaphuman}等多个数据集。接着我们还使用了SQuAD \cite{rajpurkar-etal-2016-squad}， NarrativeQA \cite{kočiský2017narrativeqareadingcomprehensionchallenge}以及RACE\cite{lai-etal-2017-race}等多个数据集来探索我们方法的潜力，它们能够探索我们的方法在长文本数据中，在文本生成和"needle in a haystack"任务的表现情况。

\subsection{Evaluation Metrics} 
We rigorously adhere to the evaluation protocols established in the original papers for each dataset. For GLUE and SuperGLUE benchmarks, we primarily employ accuracy as the standard metric. For question-answering performance on SQuAD, we utilize both F1 and Exact Match (EM) scores. The evaluation of long-form text generation in NarrativeQA follows the ROUGE-L metric, while classification performance on RACE is measured using accuracy. These metrics were selected to ensure comprehensive assessment of model capabilities across different linguistic tasks and data characteristics.

\subsection{Baselines} 
We selected various models with diverse parameter sizes to comprehensively evaluate the effectiveness of our method, including DeBERTaV3-base\cite{he2021debertadecodingenhancedbertdisentangled}, RoBERTa-base, RoBERTa-large\cite{liu2019robertarobustlyoptimizedbert}, GPT-2\cite{radford2019language}, and LLaMA2-7B, LLaMA2-13B\cite{touvron2023llama2openfoundation}. These models effectively demonstrated that our method performs well across models of different scales. We selected several baseline fine-tuning techniques to compare with our method, including Fine-tuning\cite{park-lee-2021-finetuning}, Bitfit\cite{zaken2022bitfitsimpleparameterefficientfinetuning}, LoRA\cite{hu2022lora}, as well as other improved strategies of LoRA technology such as QLoRA\cite{dettmers2023qlora} and MixLoRA\cite{li2024mixloraenhancinglargelanguage}.

% \subsection{Results Analysis}

\begin{table*}[htbp]
\centering
\small
\setlength{\tabcolsep}{3pt}
\begin{tabular}{l|cccccccccc}
\hline
\multicolumn{2}{l|}{Method} & CoLA & SST-2 & MRPC & STS-B & QQP & MNLI-m & MNLI-mm & QNLI & RTE \\
\multicolumn{2}{l|}{(Trainable params)} & MCC & Acc & F1/Acc & P/S Corr & F1/Acc & Acc & Acc & Acc & Acc \\
\hline
Fine-tune & (124M) & 63.24 & 94.61 & 89.73/85.74 & 91.00 & 88.46/91.35 & 86.27 & 86.31 & 90.13 & 76.53 \\
BitFit & (102K) & 60.27 & 93.46 & 89.05/85.22 & 90.91 & 85.61/88.95 & 83.98 & 84.38 & 91.34 & 83.39 \\
LoRA & (1.3M) & 61.64 & 94.50 & 89.54/86.03 & 90.37 & 86.95/90.24 & 84.94 & 85.50 & 90.76 & 72.20 \\
MixLoRA & (5.7M) & 60.12 & 94.04 & 88.56/85.39 & 89.40 & 87.44/90.55 & 86.29 & 85.74 & 91.29 & 79.42 \\
QLoRA & (1.3M) & 58.33 & 93.81 & 87.80/83.65 & 89.93 & 85.12/88.81 & 85.26 & 85.10 & 91.84 & 83.03 \\
SSMLoRA & (1.0M) & 61.76 & 94.15 & 91.99/88.73 & 90.46 & 87.14/90.52 & 85.73 & 85.16 & 91.91 & 76.17 \\
\hline
\end{tabular}
\caption{Performance comparison of four methods based on the RoBERTa-base model across GLUE benchmark}
\label{tab:experiment-result}
\end{table*}

\subsection{Results}
We first conducted a comparative analysis on the RoBERTa-base model using various parameter-efficient fine-tuning approaches: full parameter fine-tuning, BitFit, LoRA, QLoRA, MixLoRA, and our proposed method. The performance comparison across the GLUE benchmark is presented in Table \ref{tab:experiment-result}.
% 我们首先在小模型RoBERTa-base上，分别采用fine-tune全参数微调、bitfit微调技术、LoRA微调技术以及QLoRA和MixLoRA与我们的方法在GLUE benchmark上进行对比,如表格一所示。

For all LoRA variants, including our method SSMLoRA, we set the rank $r=8$ and applied rank decomposition to multiple components including self-attention mechanisms, feed-forward layers, and classifiers. In our approach, we specifically implemented interval-sparse insertion for Q and V modules in the self-attention mechanism while maintaining dense insertion for other layers. Notably, SSMLoRA introduces fewer trainable parameters—less than 80\% of those introduced by LoRA—making it the most parameter-efficient among all LoRA variants we used for comparison.

Our method demonstrates robust performance across multiple datasets, achieving state-of-the-art results on MRPC and QNLI tasks, and surpassing LoRA's performance on more than half of the evaluated datasets. These results indicate that SSMLoRA exhibits particularly strong performance in NLP tasks, with notable effectiveness in sentence-pair tasks, as further validated in subsequent experiments. However, we observe certain limitations in cross-domain generalization capabilities.

To further validate SSMLoRA's effectiveness on larger-scale models, we conducted experiments on LLaMA2-7B and compared SSMLoRA with LoRA, as shown in Table \ref{tab:method-comparison}. The results demonstrate SSMLoRA's significant advantages with fewer trainable parameters. We extended our evaluation to include LoRA on LLaMA2-13B, where SSMLoRA continued to show exceptional performance.

% 我们将LoRA技术及其所有变种，包括我们的SSMLoRA都设置为r=8，将秩分解应用于模型的自注意力机制，全连接层，分类器等多个位置。

% 对于我们的技术，在自注意力机制模块，只对Q和V模块采用间隔稀疏插入，其它层正常密集插入。可以看到SSMLoRA技术引入的可训练参数不足LoRA技术引入的80\%，也是LoRA所有变体中参数量最少的一种。我们的技术在多个数据集上均有较好的表现，其中在MRPC、QNLI中表现为四种微调方法的最佳，在超过半数的数据集中均表现超过LoRA微调技术。我们可以得到结论SSMLoRA技术在NLP领域的任务中有较好的表现，特别擅长于解决基于句子对的任务，这在后续的实验中也能进一步看到。但是在跨领域的泛化能力上有所不足，存在一定的短板。

% 我们进一步探索在更大的模型上验证SSMLoRA的表现结果。我们在LLaMA2-7B的模型上分别采用我们的SSMLoRA和LoRA微调技术，如表格二所示，可以看到SSMLoRA在较少训练参数的情况下有明显的优势，因此我们进一步在LLaMA2-13B的模型上采用LoRA技术进行训练，最终我们看到SSMLoRA具有非常好的结果。

\begin{table*}[htbp]
\centering
\small
\begin{tabular}{l|cccccc}
\hline
Method & RTE & BoolQ & WSC & WiC & MultiRC & COPA \\
(Trainable params) & Acc & Acc & Acc & Acc & F1 & Acc \\
\hline
SSMLoRA (15.8M,7B)) & 88.5 & 85.4 & 63.5 & 75.9 & 88.2 & 91.0 \\
LoRA (20.0M,7B)) & 85.9 & 85.2 & 64.4 & 65.5 & 84.8 & 87.0 \\
LoRA (31.3M,13B)) & 89.9 & 87.1 & 63.5 & 69.9 & 87.1 & 92.0 \\
\hline
\end{tabular}
\caption{Performance comparison of SSMLoRA and LoRA across various NLP tasks}
\label{tab:method-comparison}
\end{table*}

Based on these comprehensive evaluations, we conclude that SSMLoRA serves as an effective parameter-efficient alternative to traditional LoRA fine-tuning. Drawing inspiration from state-space models, our innovative approach incorporating state-space equations demonstrates particular strengths in specific datasets. We anticipate that these advantages will manifest in superior performance across various task domains.
% 通过这个部分，我们可以得到结论，SSMLoRA能够作为一种参数高效微调的方法很好的取代LoRA微调技术发挥作用。并且我们相信作为引入状态空间方程的创新版本，受到这一类模型的优势的启发，我们认为我们的方法在一些数据集将会有明显优异的表现。
\subsection{Discussion on Long-Context Capabilities}

\begin{table}[htbp]
\centering
\setlength{\tabcolsep}{6pt} % 调整列间距
\begin{tabular}{l|c|c}
\hline
\multirow{2}{*}{Method} & \multicolumn{2}{c}{DeBERTaV3-base} \\ 
\cline{2-3}
 & SQuAD & NarrativeQA \\
\hline
Fine-tune & 85.08/69.23 & 23.05  \\
BitFit & 55.58/39.28 & 21.29  \\
LoRA & 84.63/69.27 & 24.41  \\
SSMLoRA & 84.54/69.37 & 24.10  \\
\hline
\end{tabular}
\caption{Performance comparison of methods on DeBERTaV3-base for SQuAD and NarrativeQA datasets}
\label{tab:multi-model-results}
\end{table}

\begin{table}[htbp]
\centering
\begin{tabular}{l|ccc}
\hline
Method & All & Middle & High \\
\hline
SSMLoRA & 70.1 & 71.0 & 67.37 \\
LoRA & 68.5 & 72.0 & 65.64 \\
\hline
\end{tabular}
\caption{Performance comparison of methods on RoBERTa-base for RACE dataset}
\label{tab:race-results}
\end{table}

\begin{figure}[h] % 开始插入图片
    \includegraphics[width=\linewidth]{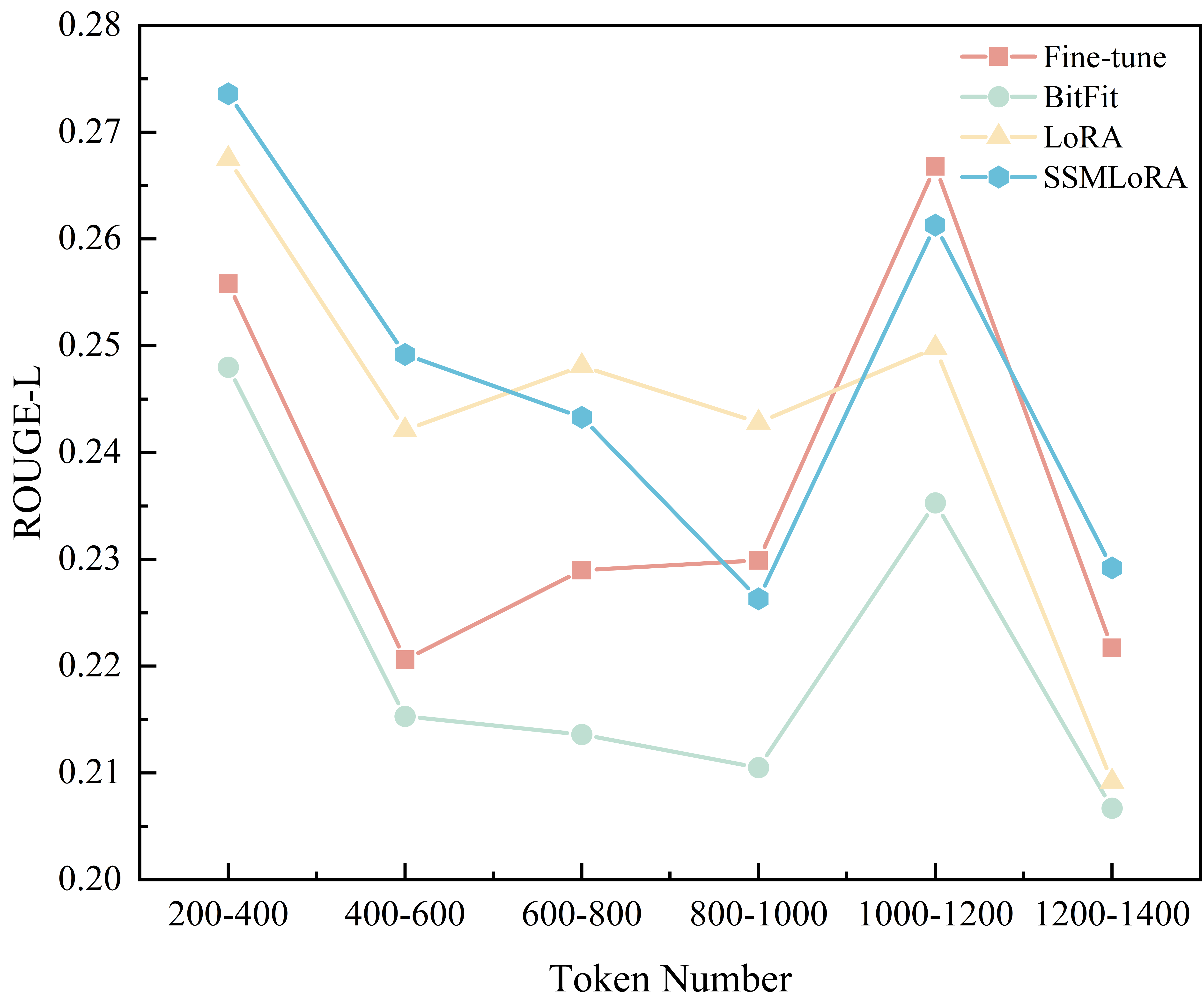} % 插入图片并设置宽度
    \caption{This figure shows the results of the DeBERTaV3-base model fine-tuned with four different techniques on the NarrativeQA dataset, evaluated on test data across various length intervals.}
    \label{fig:na}
\end{figure}

Given the architectural characteristics of State Space Models (SSMs), we aim to investigate our technique's potential in logical reasoning capabilities and long-text processing. To comprehensively explore the advantages of our method, we conduct additional evaluations on multiple datasets, demonstrating consistent improvements across different benchmarks.
%根据SSM模型的特性，我们希望探索我们的技术在逻辑推理方面的潜力以及处理长文本数据方面的潜力。为了更加全面的探索我们技术的优势，我们的还额外使用了一些其余的数据集进行评测，结果表明我们的方法均有不错的结果。

We evaluate on three datasets: SQuAD, NarrativeQA, and RACE. The SQuAD dataset requires answering questions based on Wikipedia articles, testing the model's ability to extract answers from substantial text passages, with evaluation metrics of F1 and Exact Match (EM). NarrativeQA assesses comprehension of long-form narrative structures beyond simple content matching, using ROUGE-L as its metric. The RACE dataset represents a "needle-in-a-haystack" task demanding identification of critical information within lengthy contexts. 

As shown in Table \ref{tab:multi-model-results}, we compare four fine-tuning methods on DeBERTaV3-base for SQuAD and NarrativeQA. For the RACE benchmark evaluated on RoBERTa-base, results are presented in Table \ref{tab:race-results}. Our analysis reveals that SSMLoRA achieves competitive performance on short-text matching tasks in SQuAD, demonstrating superior exact answer alignment crucial for scenarios requiring precise responses. On long-text datasets, SSMLoRA maintains strong performance, particularly excelling in the more challenging high-difficulty subset of RACE (67.37 vs. LoRA's 65.64) while showing comprehensive improvements across all difficulty levels. This suggests that the state space architecture effectively enhances logical reasoning capabilities without compromising long-text processing efficiency.

To further investigate length-dependent performance, we analyze NarrativeQA results across different text length intervals (Figure 2). The test set is partitioned into six 200-token bins based on preprocessed sequence lengths. Our method achieves state-of-the-art ROUGE-L scores in shorter contexts while demonstrating marked improvements over baseline LoRA (2.1\% relative gain) in sequences exceeding 1000 tokens, highlighting SSMLoRA's enhanced capacity for long-text comprehension.
%我们选取了SQuAD，NarrativeQA和RACE数据集，SQuAD是基于维基百科的内容提出问题的数据集，需要模型具有基于大量文本回答的能力，评估的指标是F1和EM，NarrativeQA则需要考验模型对长文本叙事结构的理解能力而不是进行简单的内容匹配，评估的指标是ROUGE-L，RACE数据集则是一个"needle-in-a-haystack" task，需要模型在较长的文本中搜寻到关键信息。我们在DeBERTaV3-base模型上使用四种方法分别微调SQuAD和NarrativeQA两种数据集，最终的结果可以在Table 3看到，在RoBERTa-base模型上采用SSMLoRA和LoRA微调RACE数据集，结果在Table 4看到。我们可以发现在短文本数据SQuAD上的匹配任务中，SSMLoRA具有能更加完美的匹配结果的能力，对于需要严格答案的场景具有更加卓越的效果。而从另外两个长文本的数据集来看，SSMLoRA具有不错的表现。特别是在RACE中，SSMLoRA在难度更高的high级别和综合所有的All上都有更好的结果。这说明了由于状态空间模型的结构优势，SSMLoRA可以显著的增强模型的逻辑推理能力，处理长文本的能力并未得到削减。
%我们进一步探索在NarrativeQA任务中不同长度的数据的表现情况。我们将NarrativeQA测试集按照预处理后的数据长度进行划分，利用微调后的模型进行测试。对于NarrativeQA数据集，具体的测试结果见Figure 2，我们将预处理后的数据按照每200个token划分为六个区间，可以看到我们的技术微调后的模型在短文本中表现优秀，ROUGE-L的值处于四种方法微调后的最高。而当token长度来到1000之后，SSMLoRA较为明显的改进了LoRA微调技术，展现出来卓越的性能。

\subsection{Discussion on Sparsity}

In this section, we explore a comparative analysis between SSMLoRA and LoRA under sparse insertion conditions, focusing specifically on their implementation within attention mechanisms—our core design component. This focused comparison allows us to examine performance differences between SSMLoRA and LoRA fine-tuning techniques when parameter disparities are amplified. Given that our introduced matrices are influenced by the scaling factor \(r\), we conducted experiments across five different values: \(r = 1, 2, 4, 8,\) and \(16\).

For our initial investigation, we selected two architectures: RoBERTa-large and GPT-2, which employ linear layers and convolutional layers in their attention modules, respectively. This selection enables us to validate SSMLoRA's effectiveness across both architectural paradigms. Our preliminary experiments on the GLUE benchmark, as shown in Tables~\ref{tab:RoBERTa-large} and~\ref{tab:gpt2}, demonstrate that SSMLoRA achieves comparable or superior performance across most tasks while utilizing only half the parameters of traditional approaches.In these experiments, SSMLoRA employs sparse insertion into the pre-trained models. For RoBERTa-large, inspired by LoRA research findings, we applied SSMLoRA exclusively to the query and value matrices within the self-attention mechanism, as concurrent modification of these matrices has been proven to yield optimal training outcomes. Regarding sparse insertion strategies, we prioritized simplicity and implementability. For RoBERTa-large, we implemented an alternating interval insertion pattern, while for GPT-2, where self-attention is encapsulated within a single convolutional layer, we modified the complete output using a skip-one interval insertion strategy. Experimental results validate the effectiveness of these design choices across multiple tasks.

To further evaluate our approach, we extended our experiments to the more challenging SuperGLUE benchmark, which presents a significant test for SSMLoRA given its reduced parameter count compared to LoRA. Maintaining consistent design strategies, we demonstrated sustained performance advantages, with detailed results presented in Table~\ref{tab:super}.

These comprehensive experiments lead to a significant conclusion: SSMLoRA, while utilizing approximately half the parameters of LoRA, achieves comparable performance across most datasets and exhibits exceptional performance on several specific tasks.

%在这个部分我们的探索了SSMLoRA在稀疏插入的情况下与LoRA技术的对比，并最后获取了实验结果，这意味着我们希望更加专注于探索仅比较在我们的核心设计——注意力机制上的插入的效果，放大参数的差异探究SSMLoRA和LoRA微调技术的效果差异。由于引入的新的矩阵受到了放缩的\(r\)的大小的影响，我们也设置了五种不同的\(r\)进行实验，分别为r=1，2，4，8，16。
%在这个部分，我们首先选取了两种模型，分别是RoBERTa-large和GPT-2，它们的注意力模块分别采用常见的线性层和卷积层组成，这同时也验证我们的SSMLoRA在线性层和卷积层均有优秀的表现。在这个部分我们先采用GLUE基准进行实验，实验的结果如\label{tab:RoBERTa-large}和\label{tab:gpt2}所示。可以看到，仅利用一半的参数就可以达到几乎相似的结果，并且在大部分任务上还有所领先。在这个部分，SSMLoRA仅采用稀疏的方式插入到预训练模型中。针对于RoBERTa-large模型，受到LoRA研究的启发，我们仅将SSMLoRA作用于自注意力机制的query和value矩阵，因为这两个矩阵被证明为同时改变它们的输出可以得到最好的训练结果。而在稀疏策略的选择上，我们致力于设计一种较为简单容易实现的方案，在RoBERTa-large上我们采用了交替间隔插入；在GPT-2上，由于自注意力机制被封装为一个完整的卷积层，我们将SSMLoRA直接改变完整的输出，在稀疏策略的设计上我们采用了间隔插入，每次都跳过一个。通过实验证明我们的设计是有效的，在多数任务上均取得了很好的结果。
%为了进一步评估，我们的实验还选取了SuperGLUE基准，一个更加困难的任务，对于仅有LoRA技术一半参数量的SSMLoRA技术是不小的挑战。我们仍然采用相同的设计策略，最终证明了效果依然很好，具体的实验结果见\label{tab:super}.

%可以得到结论，在仅有约为一半的参属下，SSMLoRA可以在大部分的数据集上均有与LoRA技术的表现，并在一些数据集上表现极为优异。

\subsection{Memory Efficiency}
While our approach achieves reduced memory requirements through its parameterization advantages, it exhibits RNN-like recurrent structures when not utilizing FFT-based optimizations. Therefore, it is crucial to thoroughly investigate the computational costs associated with our method. We present a comprehensive analysis of potential memory usage and latency during inference, parameter efficiency across pre-trained models of varying scales and training time requirements.

Our initial comparative experiments were conducted on the LLaMA2-7B model, implementing both SSMLoRA and LoRA methodologies. With a maximum sequence length of 1024 tokens, SSMLoRA consumed 31.28GB of GPU memory compared to LoRA's 31.43GB at a batch size of 4. When scaling to a batch size of 8, the memory utilization increased to 37.77GB and 38.05GB, respectively. Furthermore, we examined the memory scaling patterns of both approaches with increasing sequence lengths at a batch size of 1, as detailed in Table~\ref{tab:comparison}. The experimental results demonstrate that SSMLoRA's memory advantages become increasingly pronounced as batch sizes grow. Notably, SSMLoRA exhibits superior memory efficiency across both short and long sequence scenarios. Regarding computational efficiency, our experimental results (see Table~\ref{tab:comparison}) indicate that our proposed method introduces no significant inference latency overhead compared to LoRA when evaluated on the LLaMA2-7B model.
% 尽管我们的方法由于参数的优势会带来显存需求降低，但在不采用FFT进行特殊处理时，会出现近似于RNN的循环结构，因此很有必要进一步探究了我们的方法所带来的成本。我们将逐步探讨不同大小的预训练模型我们的方法的推理时的显存占用和延时可能，参数优势以及训练时间分析。
%我们首先在LLaMA2-7B模型上分别使用了SSMLoRA和LoRA方法进行了对比实验。在最大token数为1024的情况下，当batch size为4时，SSMLoRA和LoRA模型的显存消耗分别为31.28GB和31.43GB；当batch size增加到8时，相应的显存消耗上升到37.77GB和38.05GB。此外，我们还研究了在batch size为1的情况下，随着最大长度的增加，两种模型的内存使用规律，如附录的表格\label{tab:comparison}所示，实验结果表明，随着batch size的增加，SSMLoRA相较于LoRA的内存优势越来越明显，值得注意的是，SSMLoRA在长文和短文场景下都表现出了卓越的内存效率。至于计算效率，同样如表\label{tab:comparison}中列出的实验结果表明，在 LLaMA2-7B 模型上，与 LoRA 相比，我们提出的方法不会引入明显的推理延迟。

\subsection{Parameter Efficiency}

As previously discussed, SSMLoRA achieves parameter efficiency by incorporating sparse, rank-decomposed matrices connected through state-space models at attention mechanism positions. This approach demonstrates superior parameter reduction compared to traditional LoRA fine-tuning techniques. As illustrated in Table~\ref{tab:model-params}, the parameter reduction benefits scale proportionally with the size of the pre-trained model.
% 正如我们前面所讨论的，由于SSMLoRA能够在注意力机制的位置以稀疏的方式插入由状态空间模型连接的秩分解矩阵，因此我们的方法所需要引入的可训练参数相比LoRA微调技术有较好的降低。正如表格tab:model-params所展示的，随着预训练模型参数的增加，我们的方法减少的参数需求量也同步增加。

\subsection{Wallclock Time Efficiency
}
We conducted a comprehensive analysis of convergence times and average epoch duration across Fine-tune, BitFit, LoRA, and SSMLoRA methods, employing early stopping criteria. As demonstrated in Tables~\ref{tab:total-time} and~\ref{tab:average-time}, SSMLoRA exhibits longer total training duration and per-epoch time on smaller datasets such as CoLA, MRPC, and QNLI. However, with increasing dataset sizes, our method demonstrates significant improvements in both total training time and per-epoch duration compared to baseline Fine-tune and BitFit approaches. In comparison with LoRA, SSMLoRA requires longer training times on five datasets and increased per-epoch duration on four datasets. When considered alongside the performance metrics in Table~\ref{tab:experiment-result}, these findings suggest that SSMLoRA adopts a more gradual approach to feature learning, potentially requiring additional training epochs while ultimately achieving superior convergence results.

\section{Conclusion}
In this work, we propose a novel approach for parameter-efficient fine-tuning of large pre-trained language models, termed State Space Model Low-Rank Adaptation (SSMLoRA). While the insertion of low-rank matrices has been a common strategy, it often leads to unnecessary parameter overhead, particularly when applied uniformly across all neural network layers. To address this, we introduce an improved state space equation that strategically connects sparsely inserted low-rank matrices, significantly reducing the number of parameters while enhancing parameter efficiency. Our method is simple yet effective, and given the demonstrated potential of State Space Model, it offers a holistic improvement to LoRA techniques. Furthermore, our approach can be seamlessly integrated with existing optimization strategies, further expanding its applicability in large-scale model fine-tuning.
%我们的工作提出了State Space Model Low-Rank Adaptation (SSMLoRA)，这是一种用于参数高效微调大型预训练语言模型的创新方法。低秩矩阵的插入仍然存在很多不必要的参数浪费，我们无需在每一个神经网络层都插入。我们通过引入改良的状态空间方程连接稀疏插入的低秩矩阵，可以很好的减少参数量，提高参数的效率。我们的方法简单并且状态空间模型已被证明有很大的应用潜力，我们的方法从整体的角度来考虑LoRA技术的提高，可以很好的与现有的一些改进策略相结合。

%尽管SSMLoRA的研究取得了令人鼓舞的结果，但是我们的研究目前仍然存在一些局限性。SSMLoRA技术主要是实现了调整映射到低维度空间的输入，以此来调整输入使其能够更加适应下游的各种任务。然后这也导致了状态\( h_t \)的维度需要与映射到低秩空间的输入相匹配。例如当训练和测试采用不同的batch时，尽管能够调整\( h_t \)使其适应新的维度，但不可避免的是仍然会导致性能的下降。因此我们需要研究更好的方案来解决这个问题，防止因维度而导致的性能下降。

\section*{Limitations}
Despite the promising results achieved by SSMLoRA, our research still has some limitations. The core of SSMLoRA lies in adjusting the input mapped to a lower-dimensional space, which allows the input to better adapt to various downstream tasks. However, this also means that the state \( h_t \) must match the dimensionality of the input mapped to the low-rank space. For instance, when the batch sizes used in training and testing differ, although \( h_t \) can be adjusted to accommodate the new dimensions, this inevitably results in a performance drop. Therefore, further research is needed to develop better solutions to mitigate performance degradation caused by dimensionality mismatches.

\section*{Acknowledgments}
This work is supported by the National Natural Science Foundation of China (No. 62272092, No. 62172086), the Fundamental Research Funds for the Central Universities of China (No. N2116008) and the 18th Batch (2024) College Students' Innovative Entrepreneurial Training Plan Program of Northeastern University (No. 241264).

\bibliography{anthology,custom}
\bibliographystyle{acl_natbib}

\appendix
\section{Experimental Details}

All experiments on small-scale models, including DeBERTaV3-base, RoBERTa-base, RoBERTa-large, and GPT-2, were conducted on a single NVIDIA RTX 3090 GPU. In contrast, experiments involving larger models such as LLaMA2-7B and LLaMA2-13B were performed on a single NVIDIA RTX A6000 GPU.
%在我们的实验中，所有在小型模型，包括DeBERTaV3-base，RoBERTa-base, RoBERTa-large以及GPT-2上的实验，均在单张NVIDIA RTX 3090 GPU上完成, 而较大的模型，包括LLaMA2-7B和LLaMA2-13B上的实验，均在单张的NVIDIA RTX A6000上完成。
The initial learning rates were systematically varied within the range of $5 \times 10^{-4}$ to $1 \times 10^{-6}$. Dynamic learning rate scheduling and early stopping mechanisms were implemented to optimize training convergence.
%所有的初始学习率的范围均在5e-4到1e-6之间，然后设置了学习率的动态变化以及设置了实验过程中的早停
For all LoRA-derived techniques, we maintained consistent hyperparameters: the scaling factor $\alpha$ was set to 16, the rank was fixed at 8 (with the exception noted in Section 4.2.3), and a dropout rate of 0.1 was applied.
%对于所有LoRA的延申技术，alpha值均为16，除了4.2.3以外，rank均为8，设置了dropout为0.1.

\section{Results Supplement}

This section provides supplementary tables to the experimental results and performance analysis presented in Section 4, offering detailed empirical findings across various experiments.

%在这个部分，总共有八个表格，是section 4中实验结果部分和性能分析部分的描述的补充，是各个提到的实验的具体的实验结果。
Table~\ref{tab:rouge_l_scores_ordered}, corresponding to \label{fig:na}, presents a comprehensive analysis of model performance across different text length intervals. The table is bifurcated into two primary sections: the first demonstrates the performance of Fine-tune, BitFit, LoRA, and SSMLoRA models as token count increases, while the second section details the data volume and sampling strategy within each interval. For smaller data intervals, we conducted comprehensive testing, whereas for larger intervals, we sampled 40\% of the data. Notably, SSMLoRA demonstrates superior performance in both short and long text scenarios, consistently outperforming alternative methods. In intermediate-length texts, its performance closely approximates full-parameter fine-tuning. We attribute these observations to two primary mechanisms: for shorter texts, the sparse design mitigates overfitting and enhances textual comprehension capabilities; for longer texts, SSMLoRA leverages its state-space model advantages to overcome performance limitations inherent in other fine-tuning techniques.
%如表格Table tab:rouge_l_scores_ordered，这是\label{fig:na}的具体数值。我们完成了按照测试集数据长度划分区间，测试分别由Fine-tune，Bitfit、LoRA、SSMLoRA微调后的模型的表现结果。这个表格可以分为上下两部分，第一部分是随着数据的token量的增加，四种微调后的模型的效果；第二部分则是该数据区间的数据量，以及抽样的数量。对于数据量小的区间，我们抽取了所有数据进行测试，对于数据量大的区间，我们仅抽取了40%的数据。可以看到SSMLoRA的优势体现在文本较短和文本较长的时候，均为四种方法最佳；而在中等长度时，也有和全参数微调接近的效果。我们分析出现这样的现象基于以下的原因，对于文本较短，稀疏化的设计可以防止过拟合，可以更好的学习到理解文本的能力；对于文本较长的情况，其余微调技术的性能存在不足，但是SSMLoRA借助于其状态空间模型的优势，体现出了更好的性能。

Tables~\ref{tab:RoBERTa-large} and~\ref{tab:gpt2} utilize RoBERTa-large and GPT-2 models to evaluate SSMLoRA and LoRA performance across varying rank values \(r\) on the GLUE benchmark. Table~\ref{tab:super} extends this investigation to the SuperGLUE benchmark, with \(r\) values ranging from 1 to 16. Despite SSMLoRA employing less than 60\% of LoRA's trainable parameters, it achieves comparable performance across most datasets—an exciting validation of our parameter-efficient design.
%Table \label{tab:RoBERTa-large}和Table \label{tab:gpt2}分别使用了RoBERTa-large和GPT-2两种模型两种模型，进一步测试随着r的改变，SSMLoRA技术和LoRA技术微调后的模型在GLUE上的性能对比；Table \label{tab:super}则是进一步的探讨，我们继续使用RoBERTa-large，实验验证它们在SUPERGLUE上的性能对比。在这个部分，r的范围被设置为1，2，4，8，16。对于技术的应用方案，已在前文探讨过。对于我们可以得出，尽管SSMLoRA技术的可训练参数量不到LoRA技术的60%，但仍然做到了在大部分的数据集上具有相似的效果。我们可以得出一个十分exciting的结论，我们的技术确实做到了进一步的节约参数，实现了我们的构想。
The subsequent four tables explore computational costs. Table~\ref{tab:model-params} illustrates trainable parameters across three model sizes using full fine-tuning, LoRA, and SSMLoRA, revealing our method's increasing advantages with larger pre-trained models. Tables~\ref{tab:total-time} and~\ref{tab:average-time} analyze total training time and per-epoch duration, demonstrating SSMLoRA's superior training efficiency on larger datasets, notwithstanding marginally increased times on smaller datasets. Finally, Table~\ref{tab:comparison} documents memory and time requirements for SSMLoRA and LoRA on LLaMA2-7B across increasing token lengths at a batch size of 1. The results confirm minimal inference latency, with memory advantages becoming progressively more pronounced as batch size increases, consistent with our discussion in Section 4.3.3.
%接下来四个表格则是对我们方法的成本进行探讨。在Table \label{tab:model-params}，我们展示了三种尺寸的模型在使用全参微调、LoRA微调以及SSMLoRA微调时的可训练参数，我们的方法随着预训练模型的参数增加而显现出更为明显的优势。在Table \label{tab:total-time}，Table \label{tab:average-time}，在我们测量了四种方法的总训练时间和单个epoch的平均训练时间，通过这个部分我们发现在数据量大的数据集上SSMLoRA有更优异的训练时间，尽管在部分小数据集上出现训练时间增大，证明我们的技术在训练时间方面的优势。最后在\label{tab:comparison}上，我们记录了在LLaMA2-7B上，在batch_size为1的情况下，随着tokens长度增加，我们的方法和LoRA微调技术的显存、时间需求。可以看到推理延迟并不明显，而根据我们在4.3.3的讨论，随着batch_size的增加，显存的优势逐步明显。

\begin{table*}[t]
\centering
\begin{tabular}{|l|c|c|c|c|c|c|}
\hline
\textbf{Method} & \textbf{200-400} & \textbf{400-600} & \textbf{600-800} & \textbf{800-1000} & \textbf{1000-1200} & \textbf{1200-1400} \\
\hline
Fine-tune & 0.2558 & 0.2206 & 0.2290 & 0.2299 & 0.2668 & 0.2217 \\
BitFit    & 0.2480 & 0.2153 & 0.2136 & 0.2105 & 0.2353 & 0.2067 \\
LoRA      & 0.2675 & 0.2421 & 0.2481 & 0.2428 & 0.2498 & 0.2092 \\
SSMLoRA   & 0.2736 & 0.2492 & 0.2433 & 0.2263 & 0.2613 & 0.2292 \\
\hline
\multicolumn{7}{|l|}{\textbf{Samples in interval}} \\
\hline
Total     & 1704 & 1888 & 2334 & 3236 & 1285 & 110 \\
Evaluated & 681  & 755  & 933  & 1294 & 514  & 44  \\
\hline
\end{tabular}
\caption{The performance of the DeBERTaV3-base model fine-tuned with four different techniques on the NarrativeQA test set, evaluated across varying input sequence lengths.}
\label{tab:rouge_l_scores_ordered}
\end{table*}

%四种微调技术微调RoBERTa-base模型后，在RACE测试集不同长度数据的表现结果。

%四种微调技术微调DeBERTaV3-base模型后，在NarrativeQA测试集不同长度数据的表现结果。

\begin{table*}[h]
    \centering
    \scriptsize
    \begin{tabularx}{\textwidth}{@{}l|X|XXXXXXXXX@{}}
        \toprule
        \textbf{Method} & \textbf{\#Params} & \textbf{SST-2} & \textbf{RTE} & \textbf{CoLA} & \textbf{MRPC} & \textbf{QNLI} & \textbf{WNLI} & \textbf{QQP} & \textbf{MNLI Matched} & \textbf{MNLI Mismatched} \\ \midrule
        \textbf{LoRA} $r=1$ & 98.3K & 94.95 & 81.23 & 64.19 & 87.73/89.22 & 92.92 & 56.34 & 86.23/89.37 & 89.11 & 88.77 \\ 
        \textbf{SSMLoRA} $r=1$ & 49.2K & 95.14 & 81.23 & 64.11 & 92.15/88.97 & 93.17 & 56.34 & 85.37/89.01 & 88.37 & 87.92 \\ \midrule
        \textbf{LoRA} $r=2$ & 196K & 93.35 & 82.67 & 62.68 & 87.68/89.46 & 93.2 & 56.34 & 86.71/89.95 & 88.81 & 88.41 \\ 
        \textbf{SSMLoRA} $r=2$ & 98.5K & 95.76 & 82.67 & 63.92 & 92.39/89.46 & 93.2 & 56.34 & 85.83/89.30 & 88.68 & 88.88 \\ \midrule
        \textbf{LoRA} $r=4$ & 393K & 93.58 & 85.56 & 61.62 & 87.12/88.73 & 93.45 & 54.93 & 86.80/89.98 & 89.302 & 88.87 \\ 
        \textbf{SSMLoRA} $r=4$ & 197K & 95.76 & 85.56 & 63.79 & 93.81/91.42 & 92.24 & 66.20 & 86.55/89.85 & 88.77 & 88.86 \\ \midrule
        \textbf{LoRA} $r=8$ & 786K & 94.72 & 82.67 & 65.17 & 87.63/89.46 & 93.32 & 56.34 & 87.05/90.19 & 88.54 & 88.51 \\ 
        \textbf{SSMLoRA} $r=8$ & 396K & 95.64 & 82.67 & 65.03 & 89.95/92.61 & 92.46 & 69.01 & 79.74/85.72 & 88.75 & 88.72 \\ \midrule
        \textbf{LoRA} $r=16$ & 1.6M & 95.07 & 83.03 & 62.88 & 88.71/90.20 & 93.4 & 69.01 & 87.09/90.34 & 88.24 & 88.69 \\ 
        \textbf{SSMLoRA} $r=16$ & 798K & 95.30 & 83.03 & 65.33 & 92.93/90.20 & 93.43 & 56.34 & 81.98/86.82 & 88.91 & 88.36 \\ \bottomrule
    \end{tabularx}
    \caption{Comparison of LoRA and SSMLoRA methods on the GLUE dataset (Rank r=1,2,4,8,16), modle:RoBERTa-large}
    \label{tab:RoBERTa-large}
\end{table*}

\begin{table*}[h]
    \centering
    \scriptsize
    \begin{tabularx}{\textwidth}{@{}l|X|XXXXXXXXX@{}}
        \toprule
        \textbf{Method} & \textbf{\#Params} & \textbf{SST-2} & \textbf{RTE} & \textbf{CoLA} & \textbf{MRPC} & \textbf{QNLI} & \textbf{WNLI} & \textbf{QQP} & \textbf{MNLI Matched} & \textbf{MNLI Mismatched} \\ \midrule
        \textbf{LoRA} $r=1$ & 37.6K & 86.01 & 57.4 & 5.3 & 78.43 & 81.38 & 56.34 & 80.77/85.17 & 73.46 & 73.53   \\ 
        \textbf{SSMLoRA} $r=1$ & 18.4K & 84.86 & 64.62 & 5.54 & 78.19 & 81.51 & 57.75 & 85.37/89.01 & 75.11 & 75.32 \\ \midrule
        \textbf{LoRA} $r=2$ & 75.3K & 85.78 & 59.21 & 2.56 & 76.72 & 82.23 & 49.3 & 80.89/85.15 & 76.24 & 76.36 \\ 
        \textbf{SSMLoRA} $r=2$ & 36.9K & 86.24 & 61.37 & 5.17 & 72.79 & 82.59 & 50.7 & 85.84/89.30 & 75.51 & 75.24 \\ \midrule
        \textbf{LoRA} $r=4$ & 150K & 86.81 & 59.21 & 2.56 & 75.98 & 81.27 & 53.52 & 81.80/86.15 & 74.73 & 75.31 \\ 
        \textbf{SSMLoRA} $r=4$ & 73.9K & 86.93 & 58.48 & 6 & 76.72 & 82.23 & 40.85 & 86.56/89.86 & 76.28 & 76.61 \\ \midrule
        \textbf{LoRA} $r=8$ & 301K & 86.58 & 61.01 & 6.63 & 78.68 & 82.08 & 56.34 & 81.99/86.01 & 77.21 & 77.52 \\ 
        \textbf{SSMLoRA} $r=8$ & 148K & 87.04 & 62.45 & 4.43 & 77.94 & 83.12 & 56.34 & 81.15/85.14 & 77.04 & 77.57 \\ \midrule
        \textbf{LoRA} $r=16$ & 602k & 87.50 & 58.84 & 8.16 & 76.23 & 82.70 & 47.89 & 82.89/87.27 & 77.34 & 78.40 \\ 
        \textbf{SSMLoRA} $r=16$ & 297k & 86.58 & 59.57 & 1.81 & 76.47 & 83.95 & 43.66 & 82.0/86.2 & 78.11 & 78.50 \\ \bottomrule
    \end{tabularx}
    \caption{Comparison of LoRA and SSMLoRA methods on the GLUE dataset (Rank r=1,2,4,8,16), model: GPT-2}
    \label{tab:gpt2}
\end{table*}

\begin{table*}[h]
    \centering
    \scriptsize
    \begin{tabularx}{\textwidth}{@{}l|X|XXXXXXXX@{}}
        \toprule
        \textbf{Method} & \textbf{\#Params} & \textbf{WSC} & \textbf{WiC} & \textbf{RTE} & \textbf{BoolQ} & \textbf{COPA} & \textbf{ReCoRD (F1/Acc)} & \textbf{MultiRC (F1a/EM)} \\ \midrule
        \textbf{LoRA} $r=1$ & 98.3K & 64.42 & 71.63 & 81.95 & 80.98 & 91.0 & 23.30/57.16 & 78.45/21.95 \\ 
        \textbf{SSMLoRA} $r=1$ & 49.2K & 63.46 & 69.44 & 81.59 & 77.77 & 86.0 & 23.35/60.88 & 77.77/20.79 \\ \midrule
        \textbf{LoRA} $r=2$ & 196K & 63.46 & 70.69 & 82.31 & 81.25 & 94.0 & 23.75/58.29 & 79.30/25.25 \\ 
        \textbf{SSMLoRA} $r=2$ & 98.5K & 63.46 & 69.44 & 81.59 & 79.69 & 89.0 & 24.39/56.81 & 78.19/21.95 \\ \midrule
        \textbf{LoRA} $r=4$ & 393K & 63.46 & 69.44 & 84.48 & 62.17 & 88.0 & 24.74/56.65 & 79.28/23.93 \\ 
        \textbf{SSMLoRA} $r=4$ & 197K & 64.42 & 71.00 & 83.75 & 82.08 & 90.0 & 24.82/56.07 & 78.33/19.64 \\ \midrule
        \textbf{LoRA} $r=8$ & 786K & 63.46 & 71.00 & 81.95 & 62.17 & 84.0 & 27.01/55.79 & 76.11/20.30 \\ 
        \textbf{SSMLoRA} $r=8$ & 396K & 63.46 & 72.88 & 83.75 & 80.76 & 92.0 & 24.05/57.27 & 79.94/25.08 \\ \midrule
        \textbf{LoRA} $r=16$ & 1.6M & 63.46 & 63.64 & 83.03 & 62.17 & 94.0 & 25.30/54.93 & 80.08/23.43 \\ 
        \textbf{SSMLoRA} $r=16$ & 798K & 63.46 & 71.63 & 84.12 & 62.17 & 91.0 & 23.83/57.70 & 78.22/20.79 \\ \bottomrule
    \end{tabularx}
    \caption{Comparison of LoRA and SSMLoRA methods on the SuperGLUE dataset (Rank $r=1,2,4,8,16$)}
    \label{tab:super}
\end{table*}

\begin{table*}[htbp]
\centering
\begin{tabular}{|l|c|c|c|}
\hline
Model & RoBERTa-base & LLaMA2-7B & LLaMA2-13B \\
\hline
Base & 124M & 6.6B & 13B \\
LoRA & 1.3M & 20.0M & 31.3M \\
SSMLoRA & 1.0M & 15.8M & 24.81M \\
\hline
\end{tabular}
\caption{Parameter comparison across different model scales}
\label{tab:model-params}
\end{table*}

\begin{table*}[htbp]
\centering
\setlength{\tabcolsep}{3pt}
\begin{tabular}{l|ccccccccc}
\hline
\multicolumn{2}{l|}{Method} & CoLA & SST-2 & MRPC & STS-B & QQP & MNLI & QNLI & RTE \\
\multicolumn{2}{l|}{(Trainable params)} & time & time & time & time & time & time & time & time \\
\hline
Fine-tune & (124M) & 4.26min & 28.56min & 3.51min & 5.1min & 335.25min & 191.7min & 46.00min & 1.46min \\
BitFit    & (102K) & 2.23min & 24.82min & 1.37min & 5.03min & 511.64min & 360.37min & 142.8min & 3.79min \\
LoRA      & (1.3M) & 5.27min & 34.1min & 2.45min & 3.47min & 191.39min & 229.41min & 34.69min & 1.99min \\
SSMLoRA   & (1.0M) & 8.01min & 23.89min & 3.68min & 3.86min & 192.15min & 223.55min & 72.30min & 1.85min \\
\hline
\end{tabular}
\caption{Performance comparison of four methods based on the RoBERTa-base model across various tasks}
\label{tab:total-time}
\end{table*}

\begin{table*}[htbp]
\centering
\setlength{\tabcolsep}{3pt}
\begin{tabular}{l|ccccccccc}
\hline
\multicolumn{2}{l|}{Method} & CoLA & SST-2 & MRPC & STS-B & QQP & MNLI & QNLI & RTE \\
\multicolumn{2}{l|}{(Trainable params)} & time & time & time & time & time & time & time & time \\
\hline
Fine-tune & (124M) & 0.47min & 4.08min & 0.39min & 0.34min & 37.25min & 47.93min & 9.2min & 0.16min \\
BitFit    & (102K) & 0.32min & 3.17min & 0.15min & 0.63min & 36.55min & 40.04min & 8.93min & 0.27min \\
LoRA      & (1.3M) & 0.48min & 6.82min & 0.22min & 0.35min & 19.14min & 45.88min & 5.78min & 0.17min \\
SSMLoRA   & (1.0M) & 0.62min & 5.97min & 0.26min & 0.30min & 16.01min & 44.71min & 12.05min & 0.20min \\
\hline
\end{tabular}
\caption{Performance comparison of four methods based on the RoBERTa-base model across various tasks}
\label{tab:average-time}
\end{table*}

\setlength{\tabcolsep}{3pt}
\begin{table*}[htbp]
    \centering
    \begin{tabular}{l|r|r|r|r|r|r|r|r|r|r|r|r}
        \hline
        \textbf{max\_length} & \textbf{16} & \textbf{32} & \textbf{64} & \textbf{128} & \textbf{256} & \textbf{512} & \textbf{1024} & \textbf{2048} & \textbf{4096} & \textbf{5000} & \textbf{6000} & \textbf{7000} \\
        \hline
        \multicolumn{13}{c}{\textbf{Memory Usage (GB)}} \\
        \hline
        SSMLoRA & 24.78 & 24.79 & 24.83 & 24.89 & 25.01 & 25.30 & 25.80 & 26.81 & 28.81 & 29.69 & 30.67 & 31.65 \\
        \hline
        LoRA & 24.84 & 24.85 & 24.88 & 24.95 & 25.07 & 25.32 & 25.82 & 26.82 & 28.82 & 29.70 & 30.68 & 31.66 \\
        \hline
        \multicolumn{13}{c}{\textbf{Inference Time (s)}} \\
        \hline
        SSMLoRA & 0.053 & 0.054 & 0.056 & 0.088 & 0.160 & 0.310 & 0.615 & 1.280 & 2.740 & 3.500 & 4.270 & 5.200 \\
        \hline
        LoRA & 0.036 & 0.040 & 0.042 & 0.070 & 0.130 & 0.260 & 0.510 & 1.060 & 2.210 & 2.750 & 3.610 & 4.050 \\
        \hline
    \end{tabular}
    \caption{Inference cost of LLaMA2-7B Models}
    \label{tab:comparison}
\end{table*}

\end{document}